\documentclass[11pt, a4paper, logo, internal, copyright]{googledeepmind}
\usepackage{parskip}
\usepackage{fancyvrb}
\usepackage{tcolorbox}
\usepackage{booktabs}
\usepackage[authoryear, sort&compress, round]{natbib}
\usepackage[]{xcolor} 
\title{Writing as a testbed for open ended agents}

\correspondingauthor{sgooding@google.com}

\usepackage{multirow}
\usepackage[most,skins,theorems]{}

\definecolor{darkpurple}{RGB}{102, 51, 153}   % A deep purple (example)
\definecolor{lightpurple}{RGB}{204, 153, 255}  % A lighter purple (example)

\tcbset{
 aibox/.style={
 width=\linewidth,
 top=8pt,
 bottom=4pt,
 colback=lightpurple!20!white,  % Lighter purple for the background, slightly faded with white
 colframe=darkpurple,         % Dark purple for the frame/border
 colbacktitle=darkpurple,        % Dark purple for the title background,
 center,
 }
}
\newtcolorbox{AIbox}[2][]{aibox,title=#2,#1}
\definecolor{lightblue}{rgb}{0.22,0.45,0.70}% light blue
\definecolor{forestgreen}{rgb}{0.24,0.50,0.19}% forest green

\renewcommand{\today}{03/03/2025}
\author[1]{Sian Gooding}
\author[1]{Lucia Lopez-Rivilla}
\author[1]{Edward Grefenstette}

\affil[1]{Google DeepMind}

\begin{abstract}
Open-ended tasks are particularly challenging for LLMs due to the vast solution space, demanding both expansive exploration and adaptable strategies, especially when success lacks a clear, objective definition. Writing, with its vast solution space and subjective evaluation criteria, provides a compelling testbed for studying such problems. In this paper, we investigate the potential of LLMs to act as collaborative co-writers, capable of suggesting and implementing text improvements autonomously. We analyse three prominent LLMs --- Gemini 1.5 Pro, Claude 3.5 Sonnet, and GPT-4o --- focusing on how their action diversity, human alignment, and iterative improvement capabilities impact overall performance. This work establishes a framework for benchmarking autonomous writing agents and, more broadly, highlights fundamental challenges and potential solutions for building systems capable of excelling in diverse open-ended domains.
\end{abstract}

\begin{document}
\maketitle

\section{Introduction}
\begin{quote}
    \textit{``Writing is thinking on paper.''} \\
    \hfill --- \textbf{William Zinsser}
\end{quote}
\label{sec:introduction}
Writing is a complex and iterative cognitive process. It involves self-referential mechanisms as well as multiple refinement phases for feedback integration. Its vast solution space and subjective evaluation criteria render it a compelling testbed for evaluating the capabilities of open-ended agents, particularly Large Language Models (LLMs). Currently open-ended tasks pose significant challenges for LLMs due to many possible solution spaces~\citep{si2024can}, adapting to evolving objectives~\citep{tao2024survey}, and generalising learned strategies in the absence of explicit reward functions~\citep{kirk2024understandingeffectsrlhfllm}. LLMs show promise as collaborative writing tools~\citep{10.1145/3491102.3502030, mirowski2023co}, with one mode of collaboration being the suggestion of text improvements. However, the extent to which these models can effectively navigate the open-ended nature of writing, including its self-referential and iterative aspects, remains an open question. Specifically, the relationship between action diversity, alignment with human preferences, and iterative improvement capabilities requires investigation. This paper addresses this gap by examining the performance of LLMs within the context of autonomous writing assistance. 

This paper investigates the capabilities of Large Language Models (LLMs) as open-ended agents within the domain of writing assistance. We present an in-depth analysis of three prominent models --- Gemini 1.5 Pro, Claude 3.5 Sonnet, and GPT-4o --- evaluating their ability to generate diverse actions relating to document creation and editing, align with human preferences, and iteratively refine text. Our investigation focuses on three key research questions: (1) How does the diversity of actions suggested by each model vary, and what types of changes are proposed? (2) Which actions and resulting edited texts are preferred by human editors? (3) How does the application of these actions, in batches, impact overall document quality, considering both correctness of execution and subjective judgements? By analysing the nature of suggested edits, human preferences, and the impact on document quality, we aim to understand how these models navigate the challenges of open-endedness in the context of writing assistance.

To address these questions, we employ a multi-faceted evaluation framework integrating quantitative metrics with human evaluation. We quantify action diversity, execution correctness, and document quality to benchmark the performance of these LLMs as autonomous writing agents. Our findings contribute to the understanding of fundamental challenges in building systems capable of excelling in diverse, open-ended domains and provide a robust framework for future research in the area of writing agents.

The structure of the paper is as follows. We first introduce the challenges of open-ended tasks and writing as a testbed (Section~\ref{sec:introduction}), followed by a review of related work in LLM-assisted writing (Section~\ref{sec:background}). We then explore the diversity of model-generated actions (Section~\ref{sec:action_diversity}) and assess their quality through human evaluations (Section~\ref{sec:human_judgements}). Next, we examine models' ability to filter low-quality suggestions (Section~\ref{sec:qual_filtering}) and their effectiveness in iterative text refinement (Section~\ref{sec:refining}). Finally, we discuss broader implications for open-ended AI systems and propose future directions for writing assistance tools (Section~\ref{sec:disc-conc}).

Our analysis highlights several gaps between current LLM capabilities and the requirements of open-ended text refinement:

\begin{enumerate}
    \item While models exhibit some action diversity, they tend to favour additive elaboration and reinforcement over evaluative and critical interventions. By contrast, human editors frequently engage in deletions, simplifications, and structural adjustments, reflecting a greater willingness to critique and reduce content when necessary. This pattern aligns with broader tendencies in LLM behaviour, where models are often incentivised to be sycophantic --- affirming user input rather than challenging it. Open-ended agents will require a more balanced action space, with greater capacity to evaluate and offer substantive criticisms that lead to demonstrable improvements in task outcomes.
    \item LLMs struggle to select high-quality actions as judged by human experts, often treating suboptimal and optimal interventions as equally acceptable. Writing assistance highlights the need for models to better assess the quality and impact of suggested actions, both during generation and across multi-step refinement. Incorporating self-assessment mechanisms and rationale generation --- where the model articulates the purpose of an edit --- could support higher-quality action selection.
    \item  As LLMs increasingly operate through multi-step, self-refining processes, they face the challenge of preserving alignment with document-level goals over successive revisions. Without mechanisms to stay anchored to overarching objectives, small errors and misaligned interventions accumulate over time, leading to semantic drift --- a progressive divergence from the author's intent. This problem directly interacts with (2), as errors in action evaluation compound over iterations when goal alignment is not maintained, making it harder to recover as drift increases. This pattern generalises beyond writing assistance to other open-ended tasks involving iterative improvement, wherever autonomous systems must refine their own outputs over multiple steps.
\end{enumerate}

These three components --- exploration, evaluation, and goal alignment --- form a mutually reinforcing cycle. As illustrated by the general case of assisting humans in writing documents, open-ended agents must first uncover a sufficiently rich action space (point 1, above), then evaluate which actions are likely to be effective (point 2), and finally ensure that these actions remain aligned with task goals as they are applied across iterations (point 3). We contend that developing systems capable of addressing this full cycle will be critical for advancing both LLM-based writing assistance and open-ended autonomous performance across diverse domains. 

\section{Related Work}
\label{sec:background}
LLMs are being used increasingly in open-ended tasks where solutions are not predefined and iteration is necessary. Unlike single-step text generation, these tasks require models to refine outputs over multiple cycles, balancing exploration and goal-driven specificity. Writing assistance is a prime example: effective revision involves not just generating suggestions but evaluating and integrating them in a way that improves document quality.

\subsection{LLMs for Open-Ended Tasks}
Open-ended tasks span a range of applications, from scientific discovery to creative problem-solving. In these settings, models must generate diverse possibilities while dynamically adjusting their strategies based on evolving context. Work in evolutionary and neuro-symbolic approaches~\citep{vcrepinvsek2013exploration, anderson2020neurosymbolic} has emphasised the importance of adaptive exploration, while reinforcement learning techniques~\citep{thrun1992efficient, ladosz2022exploration} have shown that self-evaluation is crucial for iterative improvement.

Despite these advances, LLMs struggle with self-directed refinement, frequently failing to filter or prioritise high-quality actions~\citep{huang2023large, kamoi2024can, panickssery2025llm}. Their ability to propose diverse edits does not always translate into effective revision, as many suggestions remain low quality or misaligned with the broader task. This gap raises fundamental questions about how LLMs navigate revision and whether their approaches resemble expert human strategies.

One open-ended task being considered is the use of LLMs for scientific discovery; however, research shows that they lack action diversity and the ability to self-evaluate~\citep{si2024llmsgeneratenovelresearch}. Analysing $4000$ generated research ideas,~\citep{si2024llmsgeneratenovelresearch} finds that LLMs quickly reach a saturation point, with the proportion of unique ideas plateauing as generation continues, suggesting a fundamental constraint on their capacity for open-ended ideation. Furthermore, while LLM-based evaluators are often used to assess idea quality, empirical results indicate that they fail to rank ideas reliably, with accuracy sometimes barely exceeding random selection. These findings highlight key challenges in adapting LLMs for autonomous scientific reasoning.

\subsection{LLMs for Writing Assistance}
The landscape of LLM-based writing assistance is fragmented, spanning various tasks, interaction paradigms, and levels of human involvement. \citet{10.1145/3613904.3642697} propose a design space for writing assistants based on task, user, technology, interaction, and ecosystem. In this section, we categorise prior work within this space, covering structured generation, human-AI collaboration, iterative revision, and the trade-offs of AI-assisted writing.

One approach to improving LLM-based writing assistance focuses on structured generation, decomposing complex writing tasks into more controllable components. \citet{huot2024agentsroomnarrativegeneration} introduce Agents' Room, a framework inspired by narrative theory that structures story generation through specialised agents handling different subtasks. Similarly, \citet{Wang_2024} present StoryVerse, which mediates between an author's high-level intent and emergent LLM-driven character behaviours, employing an abstract act-based structure to enforce narrative coherence. These approaches highlight the potential of modular systems to provide greater control over AI-assisted writing.

Beyond structured generation, human-AI collaboration plays a central role in shaping writing workflows. \citet{guo2025penpromptcreativewriters} examine how 18 professional writers incorporate LLMs into their writing processes, developing flexible relationships with AI—ranging from assistant to collaborator or muse. Writers use AI to enhance productivity and creativity while remaining cautious about preserving authenticity, ownership, and originality. Similarly, \citet{ippolito2022creativewritingaipoweredwriting} investigate the use of Wordcraft, an AI-powered writing assistant, finding that professional writers primarily employ AI for ideation and brainstorming, yet struggle with issues related to controlling style, avoiding clichéd outputs, and ensuring coherence. These findings underscore the importance of designing AI systems that support, rather than override, user intent.

AI-assisted revision has also been explored as a means of improving generated text. \citet{du-etal-2022-read} propose Read, Revise, Repeat (R3), a human-in-the-loop revision system that categorises AI-generated suggestions by revision intent (e.g., fluency, coherence, clarity, style). While initial revision rounds show high acceptance rates, these decline with successive iterations, indicating a need for improved revision mechanisms. Relatedly, \citet{chakrabarty2025aiwritingsalvagedmitigating} analyse professional writers' edits of AI-generated text, identifying recurring issues such as clichés, redundant exposition, awkward phrasing, and inconsistent tense. Their work introduces the LAMP corpus, a dataset of 1,057 AI-generated paragraphs refined through expert edits, demonstrating that while LLM-based self-editing improves text quality, human revision remains essential.

The level of scaffolding provided by AI writing assistants significantly influences the writing process. \citet{10.1145/3613904.3642134} examine how different levels of AI scaffolding (e.g., next-sentence vs. next-paragraph suggestions) affect writing quality, productivity, cognitive burden, and user satisfaction. Their findings indicate a U-shaped relationship, where higher scaffolding can enhance writing quality and efficiency but may also reduce perceived text ownership.

Collaborative writing presents additional challenges when integrating LLM-based assistance. \citet{10.1145/3544548.3581345} find that providing rationales for edits improves collaboration, making AI-generated suggestions more acceptable to human writers. However, \citet{padmakumar2024doeswritinglanguagemodels} warn that writing with a feedback-tuned LLM can lead to decreased content diversity, as multiple users incorporate similar AI-generated suggestions, potentially homogenising perspectives. This highlights the need for careful consideration of how LLM-based writing assistance can be designed to support diverse and original content.

To address these challenges, systems such as PEER~\citep{schick2022peer} aim to replicate the human writing process by generating suggestions alongside justifications, while task-specific systems such as CoEdIT~\citep{raheja2023coedittexteditingtaskspecific} allow users to specify desired text attributes for editing, enabling more targeted assistance. Future research should explore how structured task decomposition, interactive scaffolding, and adaptive revision strategies can enhance LLM-based writing assistance while preserving writer agency and diversity.

\section{Exploring the Action Space}
In the domain of writing, we define an action as a discrete modification to a document intended to improve its quality. Examples of actions include rephrasing a sentence for clarity, adding supporting evidence to strengthen an argument, or removing redundant information. Actions can vary in scope, from minor surface-level edits (e.g., fixing grammar) to substantive structural changes (e.g., reorganising paragraphs).

We analyse the diversity of actions suggested by LLMs for document improvement using a dataset of $22$ publicly available Google Documents spanning diverse genres and writing styles. These documents were selected to represent a broad spectrum of content types, including academic articles, business memos, creative writing pieces, instructional guides, and opinion essays. This diversity ensures that our analysis captures a wide range of writing challenges and improvement opportunities.

To gather improvement suggestions, we prompted the models with the following instruction:
\newpage
\begin{quote}
\begin{verbatim}
You are an AI assistant that helps users improve their documents.
You will be shown a document and your task is to come up with a
numbered list of {num_actions} diverse actions that the user
should perform in the document to improve it. Actions should be
written in the imperative form, indicating a clear, specific and
actionable instruction for how to improve the document.
{focus_sentence} Output only the list of actions, without explanations:
Here is the document:
\end{verbatim}
\end{quote}

The prompt was developed iteratively through experimentation with a prototype system. We tested multiple variations of the prompt structure across models, observing which versions generated high-quality, actionable, and diverse suggestions. The placeholders \texttt{\{num\_actions\}} and \texttt{\{focus\_sentence\}} were designed to provide flexibility, with the former allowing specification of the number of actions to generate and the latter enabling optional guidance on particular aspects of improvement (e.g., focus on grammar, style, or structure). 

The dataset for this study maximises action diversity by applying a filtering process to the generated suggestions. Using an embedding-based similarity metric, each action is mapped into a vector space, and its similarity to prior suggestions for the same document is evaluated. Actions exceeding an empirically set similarity threshold are discarded, and additional sampling continues until a predefined level of diversity is achieved. This process ensures only unique and varied actions are included, resulting in a dataset of $5,750$ actions per model across the $22$ documents.

\section{Diversity of Generated Actions}
\label{sec:action_diversity}
\begin{figure}[h]
    \centering
    % Left side: Text snippet + Embedding Image
    \begin{minipage}{0.45\textwidth}
    \label{fig:example}
        \small % Make text smaller
        \begin{tcolorbox}[
            colback=yellow!10, % Light yellow background
            colframe=black, % Black border
            sharp corners,
            width=\textwidth, % Full width of minipage
            boxrule=0.5pt,
            title=Document Snippet
        ]
        It makes sense to plan ahead for the food needed on a backpack trip. One approach is to have prepared meals for breakfast and dinner, with lunch being anytime between the two.  
        Make a menu list for each day and take it with you on the trip.  
        Stick to the menu from day-to-day so that you know each meal is covered.
        \end{tcolorbox}
        
        % Embedding image directly below text snippet
        \centering
        
        \includegraphics[width=0.9\textwidth]{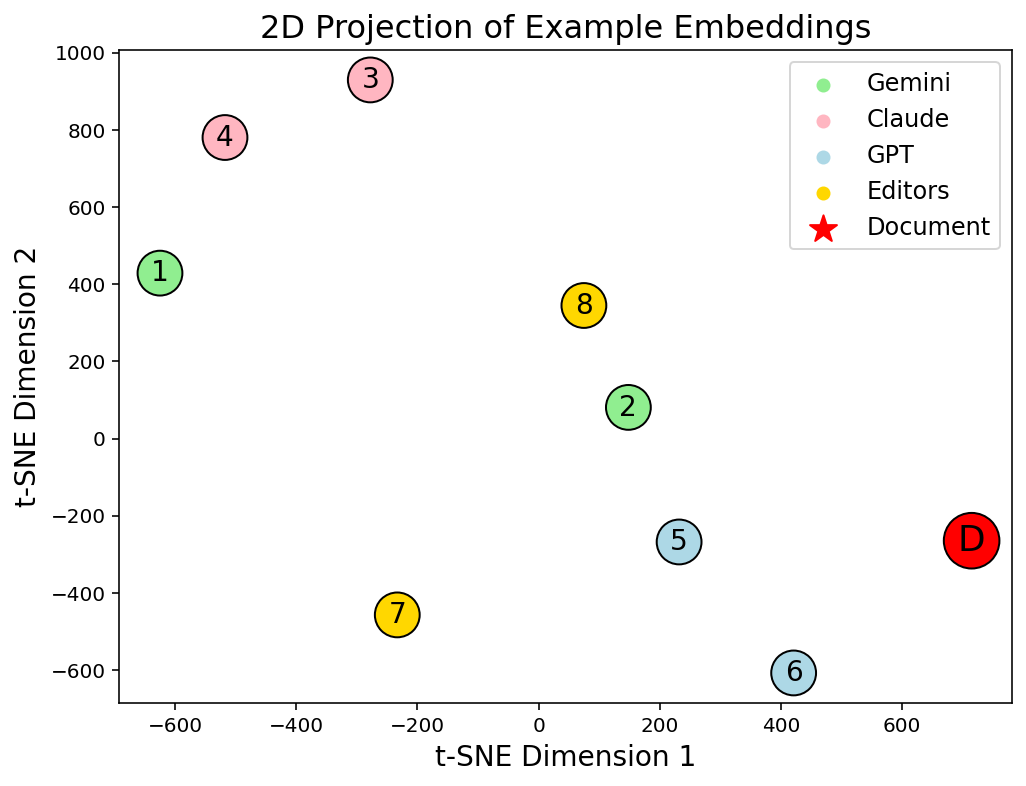} % Replace with actual file
    \end{minipage}
    \hfill
    % Right side: Table of numbered actions
    \begin{minipage}{0.5\textwidth}
        \small % Make text smaller
         
        \renewcommand{\arraystretch}{1.2} % Adjust row spacing
        \begin{tabular}{p{0.95\textwidth}} % Single column table
            \toprule
            \textbf{Gemini} \\ 
            \midrule
            (1) Proofread carefully for spelling and grammar errors. \\
            (2) Add a section on the joys and challenges of cooking and eating in the outdoors. \\
            \midrule
            \textbf{Claude} \\ 
            \midrule
            (3) Use bullet points for lists. \\
            (4) Use active voice more frequently. \\
            \midrule
            \textbf{GPT} \\ 
            \midrule
            (5) Offer portion control tips for group hiking trips. \\
            (6) Recommend specific vitamins or supplements to bring along. \\
            \midrule
            \textbf{Editors} \\ 
            \midrule
            (7) Condense the second and third paragraphs, with a heading such as: Gauging Quantity. \\
            (8) Add a title to the first section similar to ``Nutrients and Diets.'' \\
            \bottomrule
        \end{tabular}
    \end{minipage}
    % Caption for both sections
    \caption{Example excerpt from a document with randomly sampled actions from models on the right. The plot on the left visualises the embedding space of actions relative to the document $D$, where more tailored actions (e.g., 5,6) are closer to $D$, while more general actions (e.g., 1,3,4) are further away.}
    \label{fig:embedding_plot}
\end{figure}
\newpage
Action diversity is a core property of open-ended agents, as it reflects their capacity to cover a broad solution space and adapt to varied task demands. In the context of writing assistance, diverse actions enable models to propose both surface-level edits and deep structural changes. Having a limited action space constraints an agent's ability to make substantive improvements. Therefore, evaluating the diversity of model generated actions relative to human edits gives an insight into whether LLMs draw from the action space in ways that align with effective human revision strategies. 
\subsection{Action Diversity}
\begin{AIbox}{Finding 1.1}
\textit{Diversity metrics indicate significant differences in the semantic spread of actions across models, with GPT-4o achieving the highest diversity across multiple measures, while human experts outperform models on standard deviation of distances.}
\end{AIbox}

This section examines action diversity by focusing on the semantic spread of model suggested actions. To establish a benchmark for comparison, we collect annotations from expert human editors for a subset of five documents. These annotators, highly skilled in suggesting document improvements, provide an insight into the types of actions an editor would suggest. Model and human actions are then represented using Gecko embeddings~\citep{lee2024gecko}, which aim to capture nuanced semantic and intent variations. Figure~\ref{fig:embedding_plot} illustrates an example using a small subset of actions. Diversity metrics are calculated using $250$ suggestions from annotators as well as per model across these five documents and assessed using the following metrics:
\begin{itemize}
    \item \textbf{Average Pairwise Distance}: Measures the overall spread of actions in embedding space, with higher values indicating greater diversity.
    \item \textbf{Standard Deviation of Distances}: Reflects the heterogeneity in action distribution; larger values suggest more varied suggestions.
    \item \textbf{Average Centroid Distance}: Represents the average distance of actions from the centroid, where greater values indicate a broader semantic spread.
    \item \textbf{Average Nearest Neighbor Distance}: Highlights local clustering tendencies, with smaller values suggesting tighter clusters.
\end{itemize}

\begin{table}[t]
\centering
\small
\caption{Diversity metrics for model generated actions and editor suggestions (averaged over 10 runs)}
\label{tab:diversity_metrics_filtered}
\begin{tabular}{lrrrr}
\toprule
Metric & Annotator & Claude 3.5 Sonnet & Gemini 1.5 Pro & GPT-4o \\
\midrule
Avg. Pairwise Distance & $0.541 \pm 0.000$ & $\textbf{0.671} \pm 0.004$ & $0.650 \pm 0.005$ & $\textbf{0.672} \pm 0.005$ \\
Std. Dev. of Distances & $\textbf{0.140} \pm 0.000$ & $0.103 \pm 0.001$ & $0.098 \pm 0.002$ & $0.093 \pm 0.001$ \\
Avg. Centroid Distance & $0.321 \pm 0.000$ & $\textbf{0.424} \pm 0.003$ & $0.406 \pm 0.004$ & $\textbf{0.425} \pm 0.004$ \\
Avg. Nearest Neighbor Distance & $0.229 \pm 0.000$ & $0.311 \pm 0.008$ & $0.307 \pm 0.004$ & $\textbf{0.333} \pm 0.003$ \\
\bottomrule
\end{tabular}
\end{table}

As shown in Table~\ref{tab:diversity_metrics_filtered}, differences in action diversity between models are relatively small; however, across ten sampling runs, GPT-4o performs best overall. Claude 3.5 Sonnet is close behind, while Gemini 1.5 Pro shows slightly lower diversity across metrics. All models have higher diversity scores across evaluated metrics than editors, with the exception of standard deviation of distances, where human editors exhibit the highest variability. This difference is expected, as model-generated actions are sampled explicitly to maximise diversity, whereas human annotators provide edits without such constraints. Thus, the inclusion of human annotations serves as a reference point rather than a direct comparison. The higher standard deviation observed in human edits suggests that while their actions may not be as broadly distributed in embedding space, they exhibit greater flexibility in scale and type, reflecting the adaptive and context-sensitive nature of human revision strategies.

To ensure a meaningful evaluation of model capabilities, the sampling strategy for maximising model diversity was chosen because we are interested in whether high-quality suggestions can be found within the generated outputs. Without it, models tend to generate homogeneous and often superficial suggestions. By enforcing diversity, we ensure that the models explore a broader range of potential improvements rather than defaulting to common, surface-level edits. However, this also means that model outputs should be interpreted with this enforced diversity in mind, as it does not reflect the natural distribution of suggestions models would produce in an unrestricted setting. We address the impact of quality filtering in Section~\ref{sec:qual_filtering}.

\subsection{Tailoring Actions to Document Contexts}
\begin{AIbox}{Finding 1.2}
\textit{Models balance diversity and document specificity differently, reflecting an open-ended trade-off between broad exploration and goal-driven refinement. Overly diverse edits risk unfocused changes, while excessive specificity may miss broader improvements.}
\end{AIbox}

\begin{figure}[t]
    \centering
    \includegraphics[width=0.7\textwidth]{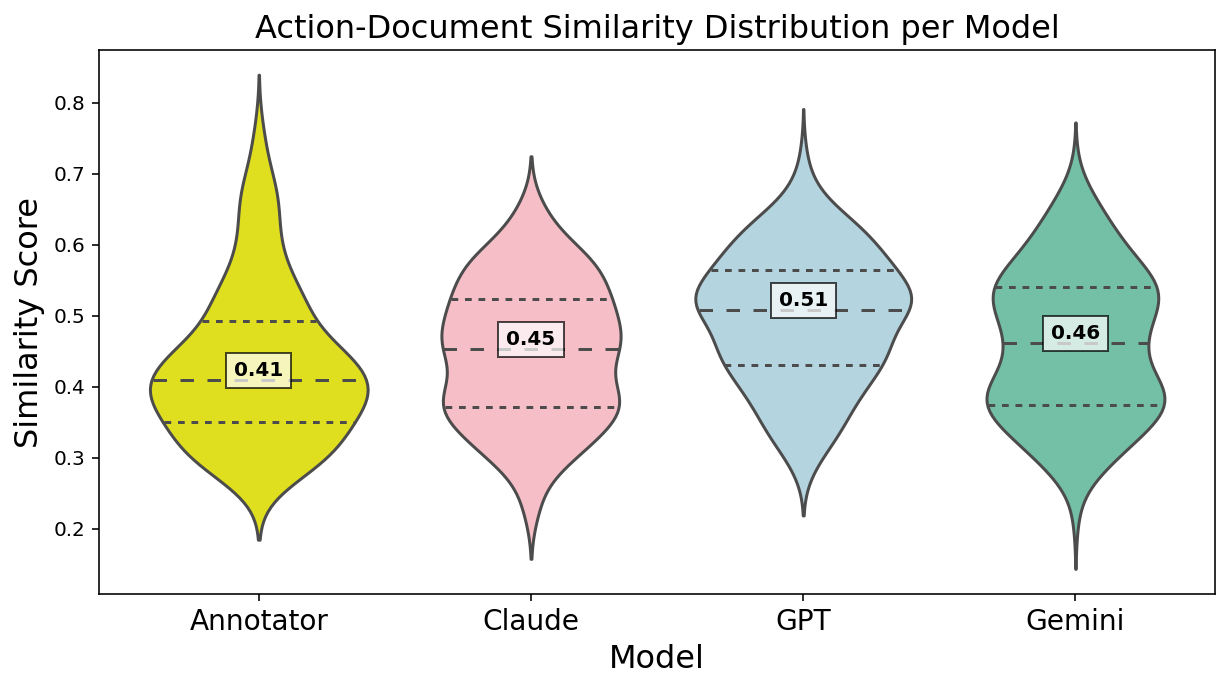}
    \caption{This violin plot shows a comparison of the models on action-document similarity, measured using cosine distance. The document and actions are embedded in the same space, and similarity is computed as the cosine distance between the document and each action. Higher similarity values indicate actions that closely align with the original document, while lower values suggest more general or diverse modifications.
}
    \label{fig:action_similarity_to_documents}
\end{figure}
The \textbf{average intra-document action similarity} measures how similar the suggested actions are to each other within the same document. Lower values indicate a wider range of action types per document, while higher values suggest more uniformity in the types of edits proposed.  

Figure~\ref{fig:action_similarity_to_documents} shows the \textbf{distribution of action-document similarity scores}, representing how closely each suggested action aligns with the original document based on cosine similarity to the document embedding. Higher values indicate actions that are more tailored to the document, while lower values reflect more general suggestions.

Our key findings are as follows:
\begin{itemize}
    \item \textbf{GPT-4o produces the most document-specific suggestions.} Figure~\ref{fig:action_similarity_to_documents} shows that GPT-4o has the highest action-document similarity, indicating that its suggested actions are the most aligned with the document’s original content.
    
    \item \textbf{Gemini 1.5 Pro introduces the greatest diversity of edits within a document.} It has the lowest intra-document similarity (0.401), meaning its suggested actions vary more within a single document compared to other models.
    
    \item \textbf{Claude 3.5 Sonnet balances diversity and document alignment.} With an intra-document similarity of 0.415 and an action-document similarity of 0.448, Claude exhibits more variation in edits than GPT-4o but remains more focused on document alignment than Gemini 1.5 Pro.
    
    \item \textbf{Human annotators maintain a consistent revision style within documents while introducing the most variation across documents.} They exhibit the highest intra-document similarity (0.451), suggesting a stable, structured approach within a document. At the same time, they have the lowest action-document similarity (0.430), indicating that their edits are less constrained by the document’s original content and thus more general.
\end{itemize}

\subsection{Linguistic Patterns in Suggested Actions}
\begin{AIbox}{Finding 1.3}
\textit{Human editors emphasise content refinement and subtraction (e.g., ``remove,'' ``change,'' ``rewrite,'' ``replace''), offering a broader scope of structural improvements. In contrast, models predominantly focus on content addition and elaboration, tending to reinforce existing text rather than critically assessing its quality.}
\end{AIbox}
\begin{figure}[t]
 \centering
 \includegraphics[width=\textwidth]{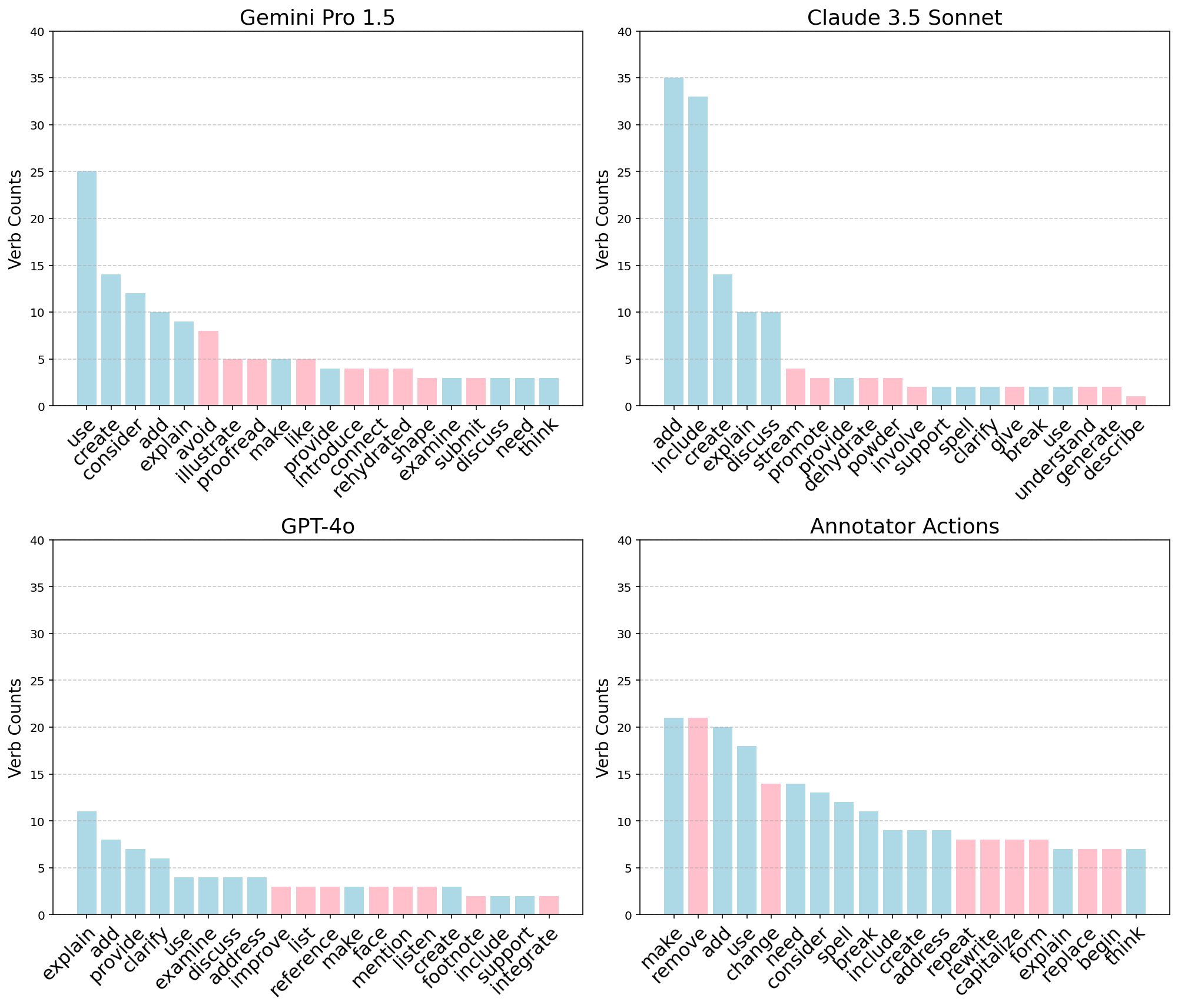}
 \caption{Comparison of the frequency of the most common verbs used in actions suggested by Gemini 1.5 Pro, Claude 3.5 Sonnet, GPT-4o, and human annotators for shared documents, highlighting commonalities and differences in their preferred improvement strategies.} 
 \label{fig:verb_counts}
\end{figure}
Figure~\ref{fig:verb_counts} displays the frequency of the most common verbs used in actions suggested by Gemini 1.5 Pro, Claude 3.5 Sonnet, GPT-4o, and human annotators. Blue bars represent verbs that appear in the top $20$ verbs for at least one other LLM or annotator set, while pink bars indicate verbs that are unique to that model or annotator within their respective top $20$. This visualisation highlights both shared patterns and distinct preferences in action verb usage, offering insights into the diversity and focus of suggested document improvements.

A significant overlap exists among the most frequent verbs across all models, with ``add,'' ``include,'' ``use,'' ``explain,'' and ``create'' consistently appearing. This shared emphasis indicates a common focus on content addition and elaboration. However, variations in the rank and frequency of specific verbs reflect distinct action preferences across models:
\begin{itemize}
    \item \textbf{Claude 3.5 Sonnet:} Exhibits a more concentrated distribution, heavily favoring verbs like ``add,'' ``include,'' and ``create,'' which  focus on content expansion. The unique verbs include document-specific terms, indicating a tailored approach to content adjustments.
    \item \textbf{Gemini 1.5 Pro:} Suggests verbs such as ``use'' and ``create'' with a relatively smoother distribution across verb categories compared to Claude. Unique verbs like ``avoid,'' ``illustrate,'' and ``proofread'' highlight a focus on enhancing content quality and clarity.
    \item \textbf{GPT-4o:} Displays a mix of content addition and structural refinement actions, with a more even distribution of verbs. Unique verbs include ``improve,'' ``list,'' and ``reference''.
\end{itemize}

Human annotators suggest actions with a more balanced verb distribution, with a higher overall usage of verbs compared to the models. Unlike LLMs, which predominantly focus on content addition and elaboration, human annotators frequently identify opportunities to ``remove,'' ``change,'' ``rewrite,'' and ``replace'' existing content. This highlights a distinction: human annotators actively engage in reshaping the text, recognising when content subtraction enhances clarity, coherence, or relevance.

Moreover, human annotators are more likely to critique the text, whereas models exhibit sycophantic tendencies~\citep{sharma2023towards}, often reinforcing or elaborating on existing content rather than challenging its quality. This bias towards positivity may result in missed opportunities for improving conciseness and addressing potential weaknesses in the original text.

\subsection{Distribution of Direct Feedback in Text}
\begin{figure}[t]
 \centering
 \begin{minipage}{0.58\textwidth} % Adjust width as needed
  \includegraphics[width=\linewidth]{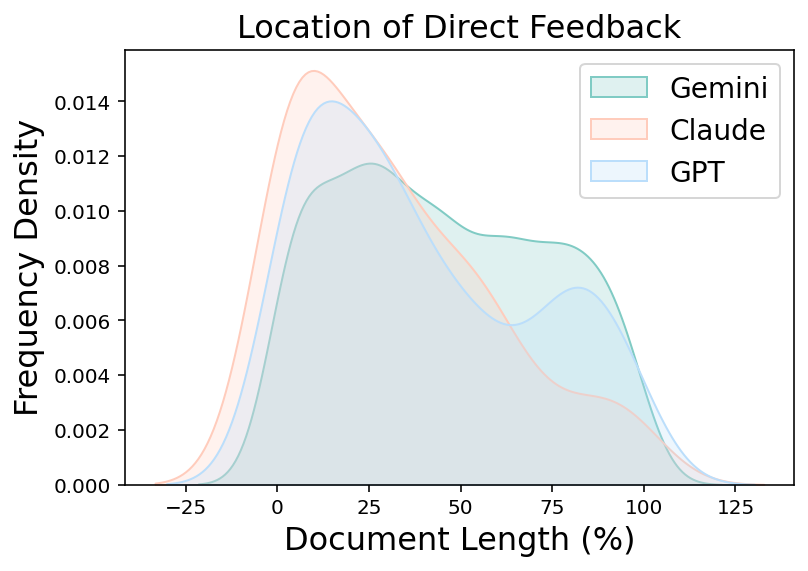} 
 \end{minipage}
 \hfill % Horizontal space
 \begin{minipage}{0.38\textwidth} % Adjust width
  {
  \begin{tabular}{lrrr}
   \toprule
   Statistic & Gemini & Claude & GPT \\
   \midrule
   Count & 1019 & 94 & 234 \\
   Mean & 45.02 & 31.35 & 39.39 \\
   Median & 42.69 & 27.21 & 31.82 \\
   Skewness & 0.18 & 0.81 & 0.50 \\
   \bottomrule
  \end{tabular}
  }
 \end{minipage}
 \caption{Analysis of Direct Feedback Location: Percentage Distribution and Descriptive Statistics Across Models} 
 \label{fig:combined}
\end{figure}
\begin{AIbox}{Finding 1.4}
\textit{Gemini 1.5 Pro provides the most direct feedback, while Claude 3.5 Sonnet and GPT-4o offer more implicit revisions. Differences in feedback placement may bias where edits are applied, compounding over refinement stages and shaping revision dynamics in open-ended writing.}
\end{AIbox}

Figure~\ref{fig:combined} examines direct feedback placement, focusing on instances where models explicitly reference portions of the text. Gemini 1.5 Pro produces significantly more direct feedback than Claude 3.5 Sonnet or GPT-4o, with a more even distribution across documents. Claude and GPT-4o, while providing tailored feedback, are less likely to quote specific text.

Overlap analysis shows some commonalities in quoted feedback between models, particularly between GPT-4o and Claude 3.5 Sonnet. However, Gemini 1.5 Pro’s feedback diverges more, suggesting a different approach in selecting and applying revisions. For example, Gemini 1.5 Pro and GPT-4o both flag \textit{``cloud storage platform''} but provide different feedback:  

\begin{tabular}{p{1.5cm}p{12cm}}
    \textbf{Gemini:} & Specify which `\textit{cloud storage platform}' you use. \\
    \textbf{GPT:} & Change `\textit{cloud storage platform}' to `\textit{cloud-based storage platform}.`
\end{tabular}

In contrast, Claude 3.5 Sonnet and GPT-4o often provide nearly identical feedback, such as defining \textit{``internet banking''} earlier in the text:  

\begin{tabular}{p{1.5cm}p{10cm}}
    \textbf{Claude:} & Define `\textit{internet banking}' early in the document. \\
    \textbf{GPT:} & Clearly define `\textit{internet banking}' in the introduction.
\end{tabular}

When considering open-ended writing, direct feedback serves as a proxy for where models apply revisions, revealing biases in both revision style and placement. As text refinement progresses through multiple stages, these biases compound, influencing both the style of feedback and where revisions are concentrated. In Section~\ref{sec:refining}, we observe that errors accumulate over time steps, suggesting that the differing feedback strategies of models --- whether favouring more localised, explicit refinements or broader, implicit changes --- can have long-term effects on the revision process.

\section{Assessing Action Quality through Human Judgements}
\label{sec:human_judgements}
\begin{AIbox}{Finding 2.0}
\textit{Gemini 1.5 Pro produces the highest quality document-improving actions as rated by annotators, significantly outperforming Claude 3.5 Sonnet and achieving a higher average than GPT-4o, though the difference is not statistically significant. This suggests that different models favour distinct revision strategies, with implications for iterative text refinement in open-ended writing.}
\end{AIbox}

To assess the quality of generated actions, we conducted human evaluations. Raters were presented with a document and a set of model-suggested changes, selecting those they deemed beneficial for improving the document. We analysed a randomly sampled set of $300$ actions ($100$ per model across Gemini 1.5 Pro, Claude 3.5 Sonnet, and GPT-4o) from five documents, with each action receiving an average of $10$ ratings. To mitigate presentation bias, the order of actions was randomised.

\begin{figure}[t]
 \centering
 \includegraphics[width=0.8\textwidth]{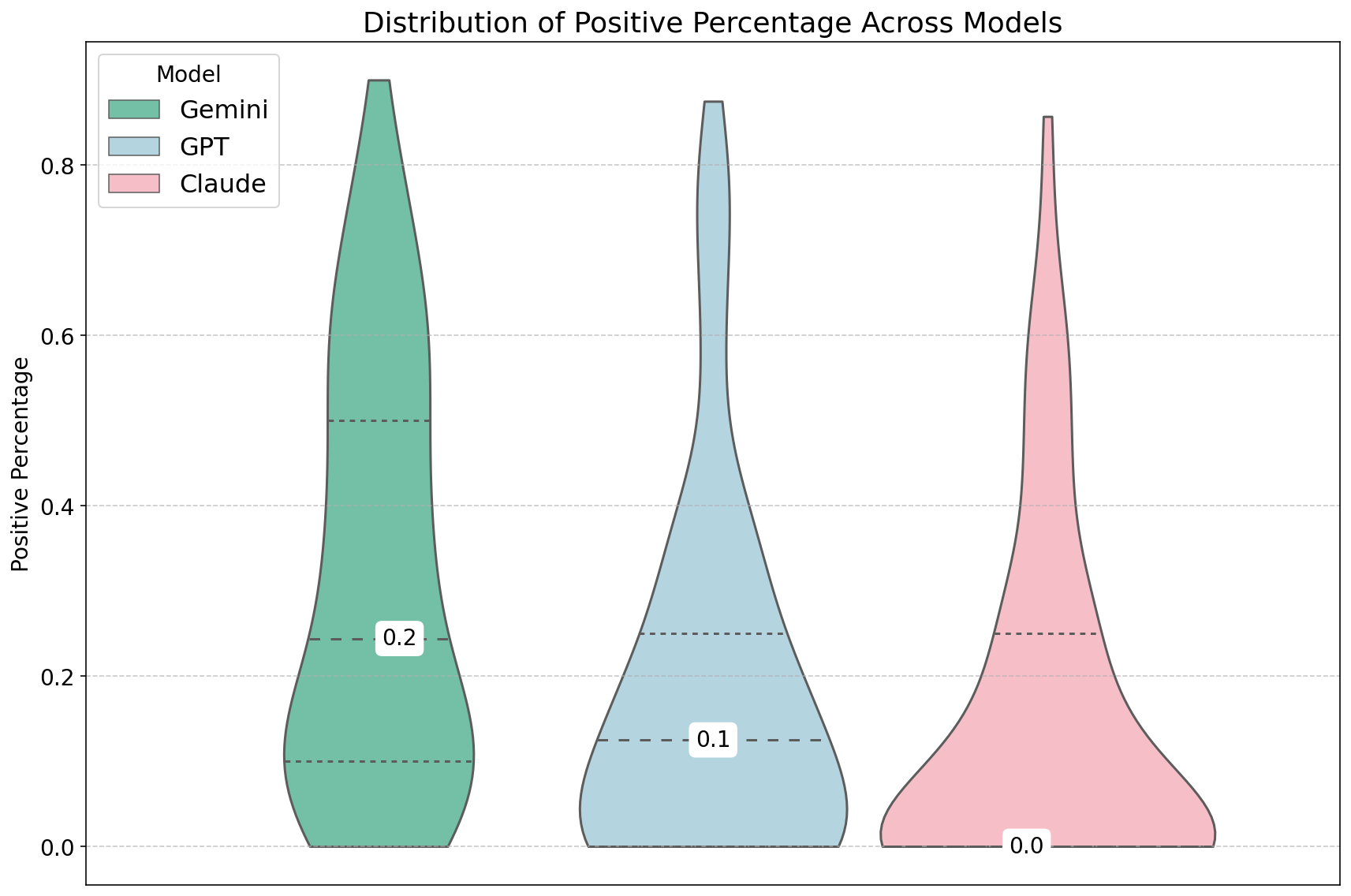}
 \caption{Violin plot showing the distribution of positive percentages for actions suggested by Gemini 1.5 Pro, GPT-4o, and Claude 3.5 Sonnet. A value of 1 indicates full agreement among $10$ annotators that the action was beneficial, while a value of $0$ means none of the annotators rated the action positively.} 
 \label{fig:suggestion_quality}
\end{figure}
\newpage
Our primary quality metric was the \textbf{positive percentage} --- the proportion of positive ratings per action. Figure~\ref{fig:suggestion_quality} shows the distribution of positive percentages for each model. Gemini 1.5 Pro achieved the highest median positive percentage, while Claude 3.5 Sonnet had the lowest, suggesting weaker overall reception. GPT-4o fell between the two, with a median closer to Claude 3.5 Sonnet than to Gemini 1.5 Pro. The distributions for GPT-4o and Claude 3.5 Sonnet were left-skewed, indicating a tendency toward lower-rated actions, whereas Gemini 1.5 Pro exhibited a more uniform distribution.

\subsection{Analysing Human Ratings with Mixed-Effects Models}
To determine whether the likelihood of selecting actions differs across models, we fit a linear mixed-effects model using Satterthwaite’s method for significance testing~\citep{kuznetsova2017lmertest}. Specifically, we modeled \texttt{num\_selected\_actions} as a function of \texttt{model}, with \texttt{document\_index} as a fixed effect and \texttt{annotator\_id} as a random effect. The model formula in R notation is:
\[
\texttt{num\_selected\_actions} \sim \texttt{model} + \texttt{document\_id} + (1|\texttt{annotator\_id})
\]
\begin{table}[t]
\centering
\caption{Fixed-effect coefficients for the number of actions selected as improving a document, by model. Random effects account for annotator variability.}
\label{tab:action_selection}
\begin{tabular}{lccc}
\hline
\textbf{Baseline Model} & \textbf{Intercept (Baseline)} & \textbf{Claude} & \textbf{Gemini} \\ 
\hline
\textbf{GPT-4o} & $5.331^{***}$ (0.479) & $-0.971^{***}$ (0.267) & $0.192$ (0.213) \\
\textbf{Gemini} & $5.523^{***}$ (0.460) & $-1.163^{***}$ (0.235) & $-0.192$ (0.213) \\
\textbf{Claude} & $4.360^{***}$ (0.491) & $0.971^{***}$ (0.267) & $1.163^{***}$ (0.235) \\
\hline
\end{tabular}
\end{table}

Table~\ref{tab:action_selection} presents the estimated fixed-effect coefficients, showing that Gemini 1.5 Pro yielded the highest average number of selected actions (5.523), followed by GPT-4o (5.331) and Claude 3.5 Sonnet (4.360). Gemini 1.5 Pro significantly outperformed Claude 3.5 Sonnet (\( p < 0.001 \)) but did not differ significantly from GPT-4o (\( p = 0.368 \)). GPT-4o also performed significantly better than Claude 3.5 Sonnet (\( p < 0.001 \)).

These results highlight Gemini 1.5 Pro's ability to generate actions that were more frequently selected as improving a document. While GPT-4o's performance was closer to Gemini 1.5 Pro than to Claude 3.5 Sonnet, it did not achieve a statistically significant advantage over Gemini. 

The random effects analysis revealed substantial variability across annotators (\( \text{Var} = 5.597 \)) compared to documents (\( \text{Var} = 0.108 \)). This suggests that different annotators had varying perspectives on what constituted a useful action, highlighting the subjective nature of revision quality. This variability leaves room for personalisation in future models, where feedback could be tailored to individual editing preferences rather than relying on a single notion of correctness.

\section{Uncovering Patterns in Human Preferences for Actions}
\begin{AIbox}{Finding 2.1}
\textit{{Clustering analysis reveals that LLM-generated suggestions fall into two broad categories: document-specific and general-purpose edits. While document-specific actions offer tailored improvements, general-purpose suggestions are widely applicable across documents.}}
\end{AIbox}

\begin{figure}[t]
    \centering
    \begin{minipage}{0.45\textwidth}
        \centering
        \includegraphics[width=\textwidth]{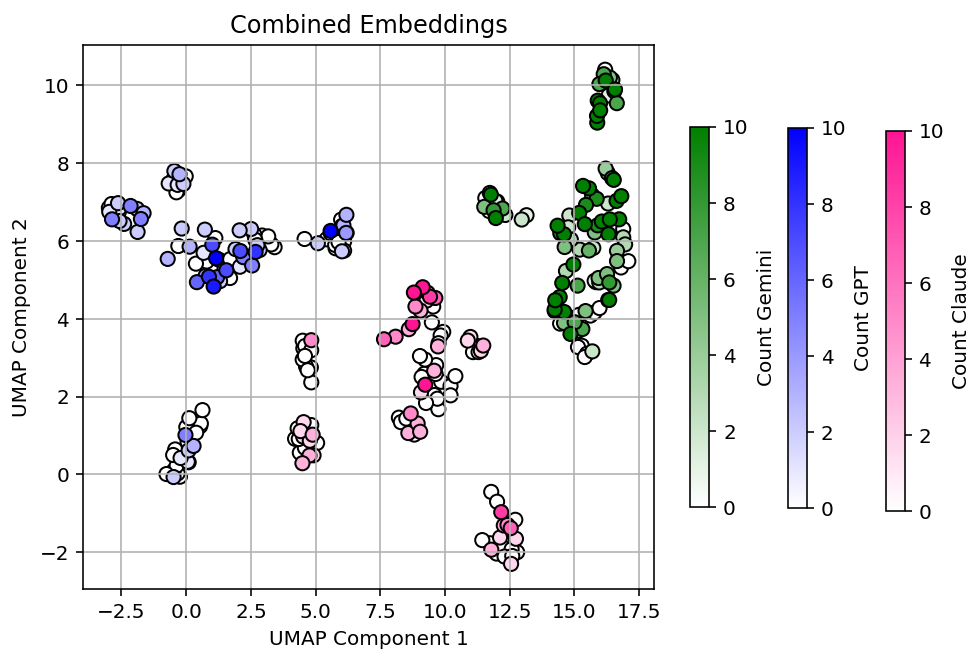}
        \caption{Clusters of actions suggested by Gemini 1.5 Pro (green), GPT-4o (blue), and Claude 3.5 Sonnet (pink), with color intensity reflecting human preference ratings. Darker shades represent higher preferences.}
        \label{fig:action_clusters}
    \end{minipage}%
    \hfill
    \begin{minipage}{0.5\textwidth}
        \centering
        \includegraphics[width=\textwidth]{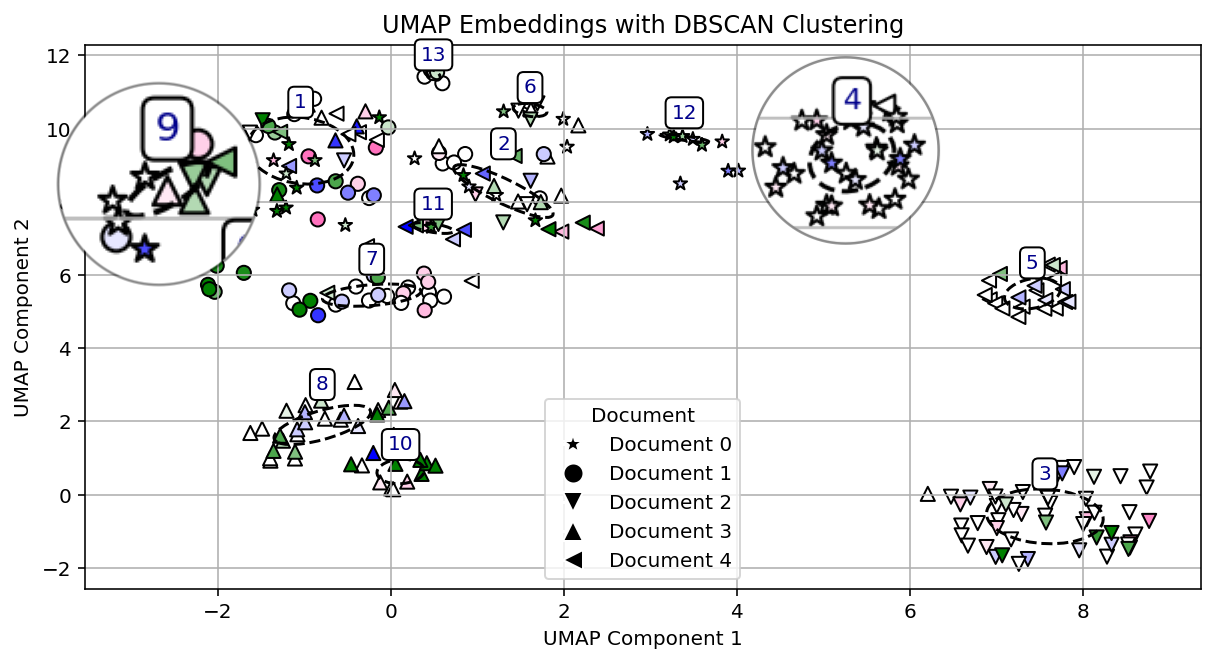}
       \caption{Clusters identified using DBSCAN based on action embeddings. The marker shapes represent the document that the action corresponds to, while the colours indicate human preferences across models, consistent with Figure~\ref{fig:action_clusters}.}

        \label{fig:dbscan_clusters}
    \end{minipage}
\end{figure}

Figure~\ref{fig:action_clusters} provides a visualisation of the semantic clustering of actions suggested by the three models, with colour intensity indicating human preference ratings. We observe that Gemini 1.5 Pro generates a greater number of highly preferred actions, as shown by the darker green clusters. Additionally, while GPT-4o (blue) and Claude 3.5 Sonnet (pink) exhibit smaller clusters, Gemini 1.5 Pro appears to produce a broader distribution of actions, leading to a larger, more generalised cluster.

When clustering actions using DBSCAN, we observe two distinct types of clusters: \textbf{document-specific clusters} and \textbf{general-purpose clusters}. Document-specific clusters consist of actions that originate from a single document, suggesting that these edits are tightly aligned with the document’s content. In contrast, general-purpose clusters contain actions spanning multiple documents, indicating that these suggestions apply more broadly to text improvement.

As shown in Figure~\ref{fig:dbscan_clusters}, \textbf{Cluster 4 is an example of a document-specific cluster}, where all actions are linked to a single document. This includes suggestions such as \textit{“Include information on music-focused cryptocurrencies”} and \textit{“Explore the role of edge computing in music streaming”}, which are directly relevant to a document on music technology. 

In contrast, \textbf{Cluster 9 represents a general-purpose cluster}, containing actions such as \textit{“Use active voice instead of passive voice”} and \textit{“Vary sentence structure to enhance readability”}. These edits appear across multiple documents, indicating their broad applicability rather than document-specific refinement.

The distinction between document-specific and general-purpose actions highlights key considerations for open-ended writing. Document-specific edits ensure deep engagement with content, while general-purpose edits introduce flexibility and consistency across revisions. In iterative refinement, models that over-prioritise general-purpose edits may produce surface-level improvements, while those focusing heavily on document-specific actions may struggle with adaptability across diverse writing contexts. 

Crucially, the effectiveness of different types of edits depends on the stage of the writing process. Early-stage drafts may benefit from document-specific feedback that refines structure and develops ideas, whereas later stages may prioritise general-purpose refinements such as grammar and clarity improvements once the core content is established. This suggests that an ideal writing assistant should adapt its feedback granularity based on the writer’s progress, offering context-aware guidance throughout the revision process.

\section{Filtering Low-Quality Suggestions with Zero-Shot and In-Context Approaches}
\label{sec:qual_filtering}
\begin{AIbox}{Finding 3.0}
\textit{Models show some ability to self-evaluate and filter low-quality actions, but performance remains limited. Gemini achieves the best filtering under moderate quality thresholds and shows slight gains from document context, while GPT-4o and Claude 3.5 Sonnet benefit more from few-shot prompting with examples. However, all models struggle to reliably align with human judgments, suggesting that effective self-evaluation in iterative writing tasks will likely require fine-tuning rather than prompting alone.}
\end{AIbox}
\begin{figure}[h]
    \centering
    \includegraphics[width=\textwidth]{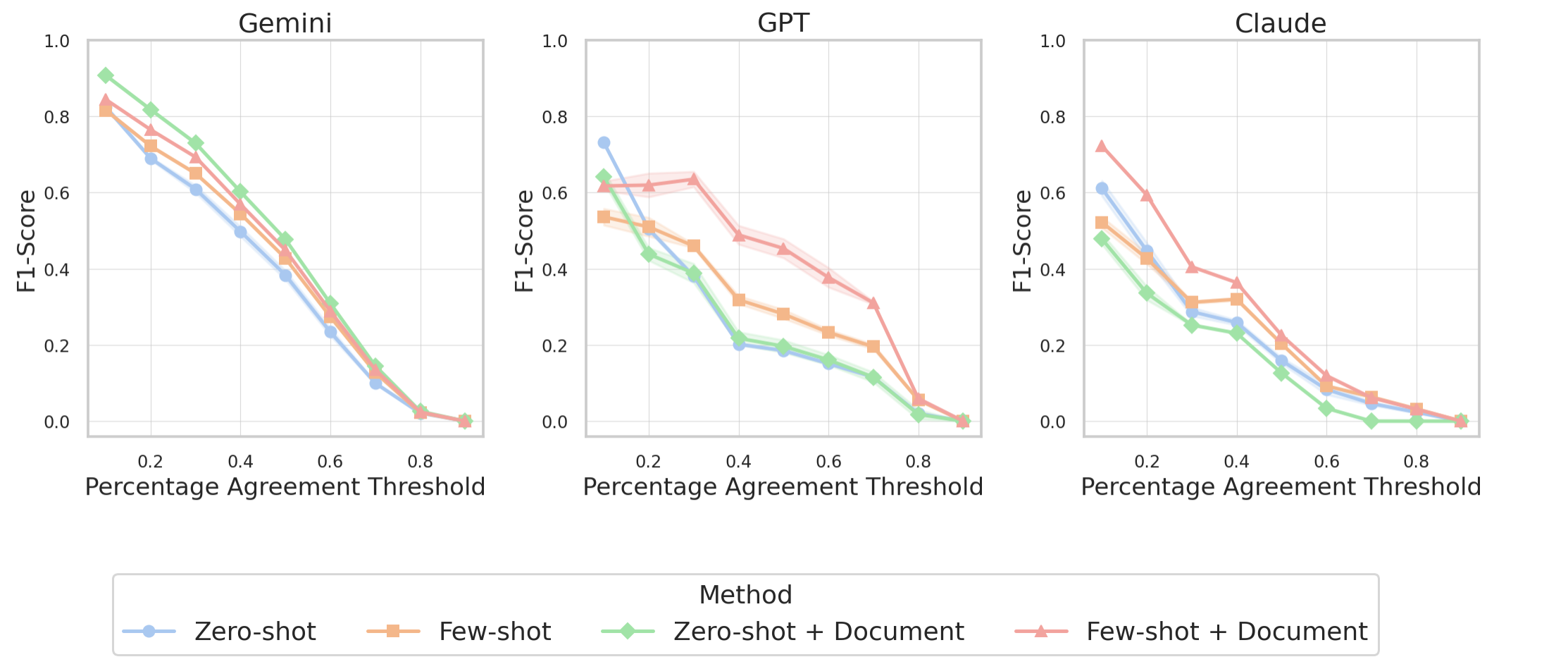}
    \caption{F1-Score Across Varying Agreement Thresholds for Different Prompting Techniques. The robustness of Gemini, GPT, and Claude's F1-score performance is shown, with results averaged over three runs of the same prompt.}
    \label{fig:f1-score}
\end{figure}
\label{sec:qual_filtering}
A core challenge for open-ended AI systems is not just producing actions but recognising which of their actions are productive. The ability to self-evaluate is critical for autonomous refinement, where models must filter high-quality outputs without human intervention. Assessing how well models can select their own best actions provides insight into the alignment between their internal quality judgments and human preferences.

To investigate this, we evaluate different prompting strategies --- zero-shot, few-shot, and document-aware prompting --- to determine how they affect a model’s ability to filter its own actions. The model is asked to assess whether each proposed action is ‘good’ (should be accepted) or not. Human ratings serve as the ground truth, and we measure performance using F1-score across increasing agreement thresholds, which require a higher proportion of annotators to agree that an action is good.

\subsection{Model Performance Across Filtering Strategies}
As expected, increasing the quality threshold reduces F1-scores across all models and prompting techniques, as stricter consensus requirements make high-confidence classifications harder.
\newpage
Beyond this, key differences emerge between models:
\begin{itemize}
    \item\textbf{GPT-4o and Claude 3.5 Sonnet perform best with few-shot prompting and examples.} Few-shot prompting with document context leads to the highest filtering performance across all thresholds, followed by few-shot without context. This suggests that these models rely more on explicit demonstrations than on document information when making acceptance decisions.
    \item \textbf{Gemini relies more on document context than examples.} 
    Unlike other models, Gemini shows little variation across prompting strategies, though we do observe a slight improvement when using zero-shot document-aware prompting compared to few-shot prompting (with or without document context). This suggests that Gemini benefits more from in-context document understanding than from few-shot example patterns.
\end{itemize}

A notable result is that, at a $0.5$ acceptance threshold---where at least half of the annotators must agree an action is good---Gemini's filtering performance is matched only by GPT-4o with \textit{few-shot and document context}, as shown in Fig.~\ref{fig:f1-score}. However, GPT-4o exhibits a substantial performance gap between its \textit{few-shot with document context} setting and its other configurations (\textit{zero-shot with document context} and \textit{few-shot alone}), whereas Gemini maintains stable, higher filtering performance across prompting variations. Claude 3.5 Sonnet, in contrast, underperforms across all evaluated settings. These findings indicate that while GPT-4o relies significantly on document context and few-shot examples to achieve optimal performance, Gemini provides robust filtering with less sensitivity to prompting strategies.

Despite these differences, absolute filtering performance remains low at the 0.5 threshold across all models. None of the prompting strategies enable models to consistently align with human quality judgments, reinforcing the idea that self-evaluation through prompting alone is insufficient. 

For open-ended AI systems that iteratively refine text, this limitation has important implications. If models cannot reliably filter their own actions, they may struggle with autonomous revision workflows, requiring human oversight to maintain quality. Fine-tuning may be necessary to improve self-selection capabilities, particularly for long-form writing tasks where revisions accumulate over multiple iterations. Without better self-evaluation, models risk amplifying low-quality refinements instead of improving document coherence and clarity.

\section{Refining Text through Successive Edits}
\label{sec:refining}
\begin{AIbox}{Finding 4.0}
\textit{Iterative text refinement depends on correctness, editing style, and user preference. While Claude 3.5 Sonnet achieves the highest correctness, its heavy insertions cause semantic drift and lower user ratings. GPT-4o, despite lower correctness, follows a conservative strategy that maintains stable improvements, while Gemini 1.5 Pro balances correctness with a more flexible approach. Across all models, excessive insertions reduce user preference, even when correct. These findings highlight a key challenge for open-ended writing: refinements must be grounded in the document’s purpose --- rigid edits can miss opportunities for improvement, while excessive changes risk distorting meaning.}
\end{AIbox}

\begin{figure}[tp]
    \centering
    % First figure - Claude
    \includegraphics[width=0.9\textwidth]{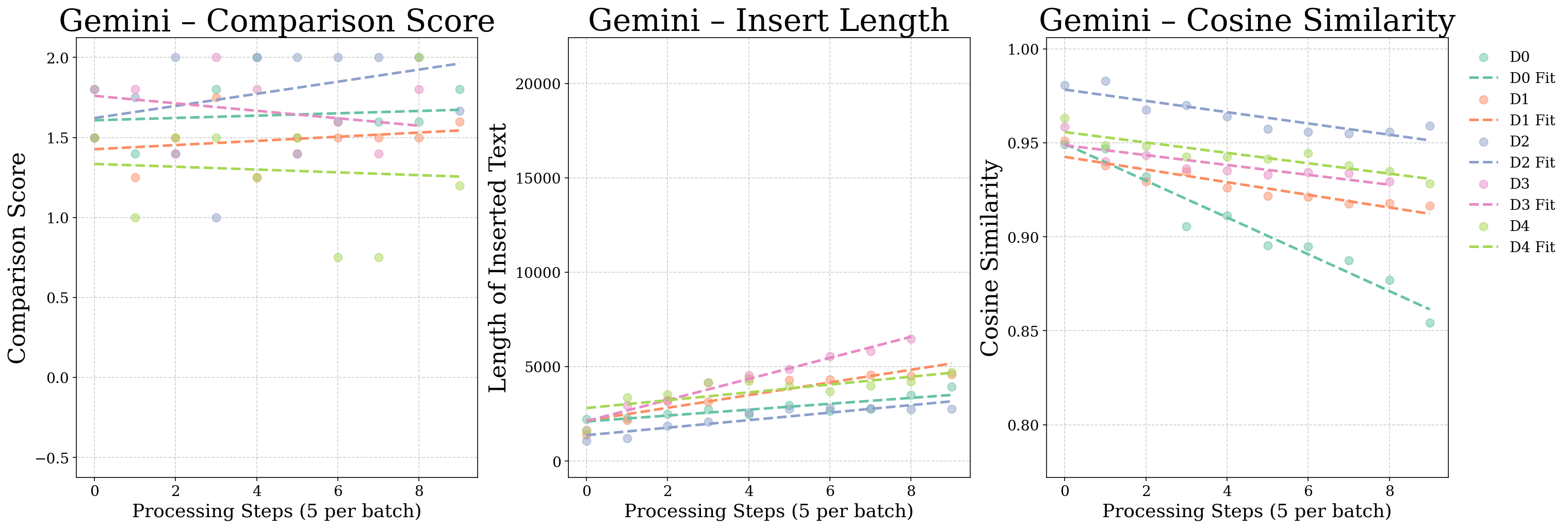}
    \vspace{0.3cm} % Space between plots
    
    % Second figure - Gemini
    \includegraphics[width=0.9\textwidth]{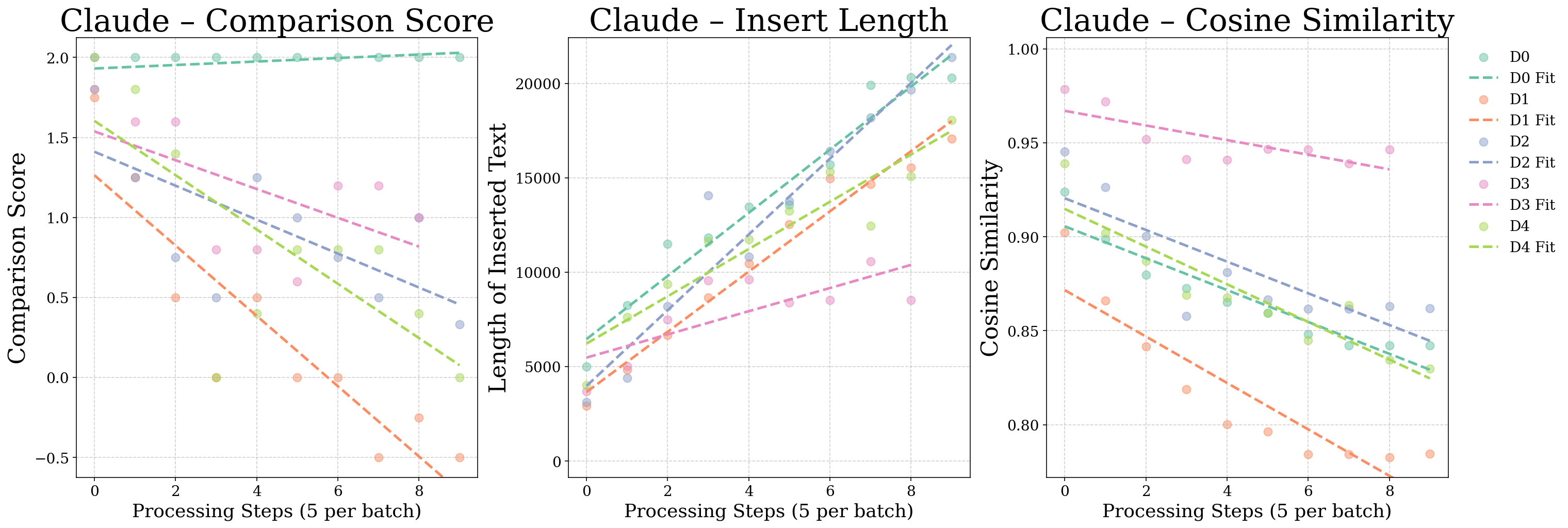}
    \vspace{0.3cm} % Space between plots
    
    % Third figure - GPT
    \includegraphics[width=0.9\textwidth]{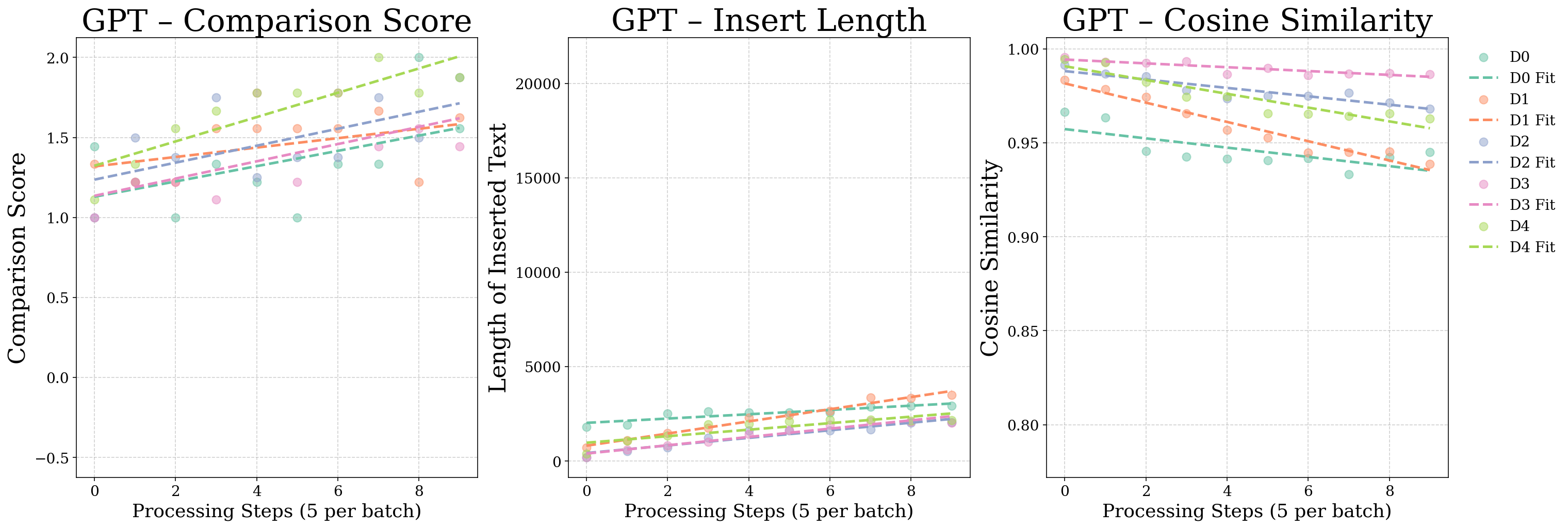}
    \small
\caption{{Evaluation of iterative refinement in LLM-generated document revisions across Gemini, Claude, and GPT, presented from top to bottom. Each plot tracks key metrics over $10$ sequential revision steps, with actions applied in batches of $5$. Best fit lines represent trends in individual sampled documents undergoing revision. The {comparison score} reflects how user preference for a document changes as edits accumulate, indicating whether revisions improve or degrade the text. {Insertion length} measures the amount of new text added at each step, capturing the extent of document expansion. {Cosine similarity} quantifies semantic shift from the original document, where lower values indicate greater divergence. These metrics reveal how effectively models integrate and refine suggested actions while preserving document quality.}}

    \label{fig:embedding_comparison}
\end{figure}

We investigate the ability of large language models to iteratively refine text by assessing whether their self-generated improvements align with human preferences. Our evaluation framework consists of two components: document comparison and action analysis.

In the \textbf{document comparison phase}, annotators compare original and revised versions of a document, rating revisions based on clarity, consistency, accuracy, and relevance. To ensure a diverse range of refinements, actions are selected by maximising their cosine distance from previously applied edits. The primary reason for sampling diverse actions is to avoid repeatedly applying the same set of edits, thus enabling exploration of a broader set of potential improvements. Actions are applied in batches of five, with document states recorded after each batch, resulting in 10 sequential revision steps per document. After each batch, a new set of candidate actions is re-sampled based on the current revised document state, ensuring subsequent edits are contextually informed by previous changes. Annotators rate quality improvement after each batch using a 5-point Likert scale, allowing us to track iterative progress across multiple revision cycles.

In the \textbf{action analysis phase}, annotators evaluate the correctness of applied actions --- assessing whether they were executed as intended.

Figure~\ref{fig:embedding_comparison} visualises document preference, insertion length, and semantic similarity over 10 revision steps across models.

\subsection{Assessing the Effectiveness of Refinement Approaches}
This section explores how LLM-generated refinements impact document quality, highlighting differences in editing strategies and their influence on correctness and user preference. Figure~\ref{fig:embedding_comparison} illustrates distinct revision strategies:
\begin{itemize}
    \item \textbf{GPT-4o produces the most stable perceived improvements.} It follows a conservative editing strategy, introducing minimal text while maintaining semantic similarity to the original document. This results in a steady increase in perceived quality.
    \item \textbf{Gemini 1.5 Pro balances correctness with a more flexible editing style.} It applies more substantial edits than GPT-4o but avoids excessive insertions, leading to a moderate improvement trajectory.
    \item \textbf{Claude 3.5 Sonnet introduces the most text, leading to greater semantic divergence.} While its edits are highly correct in execution, they often result in substantial insertions that are less favoured.
\end{itemize}
\subsubsection*{Measuring Correctness, Preferences, and Editing Styles}
We fitted linear mixed-effects models to quantify the relationships between correctness, user ratings, and editing style.

Claude 3.5 Sonnet exhibited significantly higher execution correctness than GPT-4o ($\beta = 0.481$, $p = 0.011$) and Gemini 1.5 Pro ($\beta = 0.509$, $p = 0.021$), with no significant difference between GPT-4o and Gemini ($p = 0.888$). However, despite this correctness advantage, {Claude 3.5 Sonnet received lower user ratings than both Gemini ($\beta = 0.584$, $p < 0.001$) and GPT-4o ($\beta = 0.209$, $p = 0.004$).} This suggests that while correctness is important, editing style also plays a key role in user preference.

Editing style differences were substantial:
\begin{itemize}
    \item {Claude 3.5 Sonnet inserted significantly more content} than GPT-4o ($\beta = -9560.8$, $p < 0.001$) and Gemini 1.5 Pro ($\beta = -8033.6$, $p < 0.001$).
    \item {GPT-4o made fewer deletions than Claude 3.5 Sonnet} ($\beta = -926.2$, $p < 0.001$), while Gemini 1.5 Pro exhibited deletion behavior similar to Claude ($p = 0.07$).
\end{itemize}

Importantly, {insertion volume had no significant effect on correctness ($p = 0.748$), but larger insertions correlated with lower user ratings ($p < 0.001$), while larger deletions correlated with higher ratings ($p < 0.001$).} This indicates that while correctness is crucial, excessive additions can lead to semantic drift, reducing perceived quality.

These findings highlight the importance of goal-grounded refinement in open-ended iterative tasks. While Claude 3.5 Sonnet correctly applied its suggested edits, its heavy insertions often shifted the nature of the document, leading to lower user ratings. This suggests that correctness alone is not enough --- successful refinements must align with the document’s intended purpose. Models that prioritise accuracy but disregard context risk distorting meaning, while those that preserve intent with a more measured approach see more consistent improvements. 

More broadly, this analysis reveals how different aspects of the revision pipeline --- such as action diversity and user preference --- compound across iterations, leading to distinct refinement patterns. The way models generate, filter, and apply actions not only affects individual edits but also amplifies over time, shaping the overall trajectory of document improvement.

\section{Summary of Findings and Broader Implications}

To conclude, we briefly discuss how the findings presented above in the setting of creative writing point to broader research questions in the subfield of open-ended agent research. We then provide a set of recommendations for future research directions for the production of LLM-based assistive agents for writing --- and more generally for human-facing tasks. 

\label{sec:disc-conc}
\subsection{Implications for Open-Ended Agent Development}
Our findings reveal important limitations in current LLMs that speak directly to the broader challenge of developing open-ended agents capable of adaptive, exploratory, and evaluative action. While models demonstrate high action diversity, this diversity largely reflects breadth in surface-level suggestions rather than the kind of strategic, goal-directed exploration characteristic of human editors. Moreover, a substantial proportion of model-generated suggestions are low quality and would be rejected by human editors, highlighting a lack of effective self-evaluation. This suggests that while models can generate a wide range of potential edits, they struggle to assess which refinements meaningfully contribute to document improvement.

Human experts not only exhibit greater variability in their actions but also show a particular proficiency in employing subtractive and refinement-based edits --- removing, simplifying, and restructuring content --- whereas models display a marked bias toward addition and elaboration. This suggests that LLMs, even when diverse, often operate within a narrow band of constructive, additive behaviours, limiting their capacity for critical evaluation and adaptive refinements.

This tendency reveals a broader issue with LLMs as open-ended agents: their action spaces may be wide but are not yet sufficiently flexible, responsive, or evaluative. Open-endedness requires more than generating diverse outputs; it demands the ability to expand the action space dynamically in response to context, identify when to contract or simplify, and evaluate the success of different interventions. Our observation that Gemini 1.5 Pro excels in the locality and consistency of direct feedback suggests that targeted, context-sensitive interventions are a promising direction but remain underdeveloped across models.

Crucially, these gaps point to a need for LLMs to develop more robust mechanisms for critical evaluation and action selection. The human ability to decide when to expand on an idea, when to simplify, and when to discard is deeply tied to the open-ended, iterative nature of expert problem-solving --- particularly in scientific development, where hypothesis refinement loops depend on evaluating and subtracting unproductive paths. Current models lack this evaluative loop; they can produce diversity but struggle to navigate it effectively toward task-appropriate, goal-driven outcomes.

Moving forward, progress towards open-ended LLM agents will require advances along several axes:
\begin{enumerate}
    \item Action space expansion: Training methods that encourage not only diversity but also the exploration of less frequent, high-value actions, particularly subtraction and constructive criticism.
    \item Evaluation-driven search: Mechanisms for evaluating the utility of different interventions --- both during training and inference --- to foster adaptive decision-making.
    \item Context-sensitive targeting: Attention and reasoning capabilities that enable models to identify specific points of intervention and tailor actions to local context.
    \item  Justification and rationale: Systems that can explain why an action is proposed, facilitating both human alignment and internal consistency.
\end{enumerate}

Ultimately, these capabilities are fundamental not just for writing assistance but for any domain requiring open-ended problem-solving --- from creative tasks to scientific inquiry --- where progress often depends as much on the critical analysis of ideas as on their generation.

\subsection{Future Directions for LLM-Based Writing Assistance}
The findings presented in this paper carry significant implications for the development of LLM-based writing assistants. Our investigation reveals that effective writing support extends beyond mere grammatical correctness or adherence to stylistic conventions. Instead, it hinges on a delicate balance of action diversity, alignment with human preferences, and the capacity for iterative refinement. These elements collectively determine the perceived value and practical utility of LLMs as collaborative co-writers.

Specifically, we identified the following crucial factors:
\begin{itemize}
    \item {Action diversity:} Writing assistants should offer a broad spectrum of potential improvements, encompassing not just additions and elaborations but also deletions, revisions, and structural adjustments. 
    \item {Contextual understanding:}  Effective writing assistance requires an understanding of the document's purpose, audience, and authorial intent. The tendency of some models to suggest irrelevant or inappropriate content additions highlights the challenge of ensuring that actions are contextually grounded and aligned with the overall goals of the writing task.  Localised and consistent feedback as seen in Gemini 1.5 Pro would be a promising avenue of future work.
    \item {Human preference alignment:} Ultimately, the success of a writing assistant depends on its ability to anticipate and accommodate human preferences. Our finding that raters consistently favored deletions over insertions suggests that conciseness and clarity are highly valued in the editing process.  Furthermore, the models' challenges in reliably selecting their own best actions indicate the need for improved alignment between internal quality assessments and human judgments.
    \item {Iterative refinement:}  Writing is an inherently iterative process, and LLM-based writing assistants should be designed to support this.  The observed improvements in document quality through sequential application of actions, even with imperfect execution, demonstrate the potential for LLMs to facilitate ongoing refinement and improvement.
\end{itemize}

Therefore, the development of future LLM-based writing tools should prioritise the following:
\begin{enumerate}
    \item Implement mechanisms to expand action diversity, ensuring a broad range of potential interventions beyond simple addition and elaboration. Fine-tune models to explore under-represented actions, such as deletions, revisions, and structural adjustments.
    \item A foundational requirement for future writing assistants is robust action generation and selection through improved model self-judgment. Potential solutions could be integrating rationale-driven self-assessment, forcing the model to justify feedback choices, and contextual grounding, ensuring relevance to the document's specific need and author goals.
    \item Incorporate mechanisms for understanding and responding to user feedback, allowing models' to learn and adapt to individual writing styles and preferences.
    \item Designing interfaces and workflows that facilitate iterative revision, enabling writers to integrate LLM-suggested improvements into their writing process naturally.
\end{enumerate}

In this paper, we have made the case that metric and evaluation design in open-ended creative tasks points us to the frontier of research into increasingly general human-facing agents. We believe further investment in this area, and corresponding data collection and modelling efforts are on the critical path towards artificial general intelligence. 

\section*{Acknowledgments}
We thank Laura Rimell, Piotr Mirowski, and João Madeira Araújo for their insightful feedback and suggestions, and Peter Battaglia for supporting this research.
\bibliography{main}

\begin{thebibliography}{28}
\providecommand{\natexlab}[1]{#1}
\providecommand{\url}[1]{\texttt{#1}}
\expandafter\ifx\csname urlstyle\endcsname\relax
  \providecommand{\doi}[1]{doi: #1}\else
  \providecommand{\doi}{doi: \begingroup \urlstyle{rm}\Url}\fi

\bibitem[Anderson et~al.(2020)Anderson, Verma, Dillig, and Chaudhuri]{anderson2020neurosymbolic}
G.~Anderson, A.~Verma, I.~Dillig, and S.~Chaudhuri.
\newblock Neurosymbolic reinforcement learning with formally verified exploration.
\newblock \emph{Advances in neural information processing systems}, 33:\penalty0 6172--6183, 2020.

\bibitem[Chakrabarty et~al.(2025)Chakrabarty, Laban, and Wu]{chakrabarty2025aiwritingsalvagedmitigating}
T.~Chakrabarty, P.~Laban, and C.-S. Wu.
\newblock Can ai writing be salvaged? mitigating idiosyncrasies and improving human-ai alignment in the writing process through edits, 2025.
\newblock URL \url{https://arxiv.org/abs/2409.14509}.

\bibitem[{\v{C}}repin{\v{s}}ek et~al.(2013){\v{C}}repin{\v{s}}ek, Liu, and Mernik]{vcrepinvsek2013exploration}
M.~{\v{C}}repin{\v{s}}ek, S.-H. Liu, and M.~Mernik.
\newblock Exploration and exploitation in evolutionary algorithms: A survey.
\newblock \emph{ACM computing surveys (CSUR)}, 45\penalty0 (3):\penalty0 1--33, 2013.

\bibitem[Dhillon et~al.(2024)Dhillon, Molaei, Li, Golub, Zheng, and Robert]{10.1145/3613904.3642134}
P.~S. Dhillon, S.~Molaei, J.~Li, M.~Golub, S.~Zheng, and L.~P. Robert.
\newblock Shaping human-ai collaboration: Varied scaffolding levels in co-writing with language models.
\newblock In \emph{Proceedings of the 2024 CHI Conference on Human Factors in Computing Systems}, CHI '24, New York, NY, USA, 2024. Association for Computing Machinery.
\newblock ISBN 9798400703300.
\newblock \doi{10.1145/3613904.3642134}.
\newblock URL \url{https://doi.org/10.1145/3613904.3642134}.

\bibitem[Du et~al.(2022)Du, Kim, Raheja, Kumar, and Kang]{du-etal-2022-read}
W.~Du, Z.~M. Kim, V.~Raheja, D.~Kumar, and D.~Kang.
\newblock Read, revise, repeat: A system demonstration for human-in-the-loop iterative text revision.
\newblock In T.-H.~K. Huang, V.~Raheja, D.~Kang, J.~J.~Y. Chung, D.~Gissin, M.~Lee, and K.~I. Gero, editors, \emph{Proceedings of the First Workshop on Intelligent and Interactive Writing Assistants (In2Writing 2022)}, pages 96--108, Dublin, Ireland, May 2022. Association for Computational Linguistics.
\newblock \doi{10.18653/v1/2022.in2writing-1.14}.
\newblock URL \url{https://aclanthology.org/2022.in2writing-1.14/}.

\bibitem[Guo et~al.(2025)Guo, Sathyanarayanan, Wang, Heer, and Zhang]{guo2025penpromptcreativewriters}
A.~Guo, S.~Sathyanarayanan, L.~Wang, J.~Heer, and A.~Zhang.
\newblock From pen to prompt: How creative writers integrate ai into their writing practice, 2025.
\newblock URL \url{https://arxiv.org/abs/2411.03137}.

\bibitem[Huang et~al.(2023)Huang, Chen, Mishra, Zheng, Yu, Song, and Zhou]{huang2023large}
J.~Huang, X.~Chen, S.~Mishra, H.~S. Zheng, A.~W. Yu, X.~Song, and D.~Zhou.
\newblock Large language models cannot self-correct reasoning yet.
\newblock \emph{arXiv preprint arXiv:2310.01798}, 2023.

\bibitem[Huot et~al.(2024)Huot, Amplayo, Palomaki, Jakobovits, Clark, and Lapata]{huot2024agentsroomnarrativegeneration}
F.~Huot, R.~K. Amplayo, J.~Palomaki, A.~S. Jakobovits, E.~Clark, and M.~Lapata.
\newblock Agents' room: Narrative generation through multi-step collaboration, 2024.
\newblock URL \url{https://arxiv.org/abs/2410.02603}.

\bibitem[Ippolito et~al.(2022)Ippolito, Yuan, Coenen, and Burnam]{ippolito2022creativewritingaipoweredwriting}
D.~Ippolito, A.~Yuan, A.~Coenen, and S.~Burnam.
\newblock Creative writing with an ai-powered writing assistant: Perspectives from professional writers, 2022.
\newblock URL \url{https://arxiv.org/abs/2211.05030}.

\bibitem[Kamoi et~al.(2024)Kamoi, Zhang, Zhang, Han, and Zhang]{kamoi2024can}
R.~Kamoi, Y.~Zhang, N.~Zhang, J.~Han, and R.~Zhang.
\newblock When can llms actually correct their own mistakes? a critical survey of self-correction of llms.
\newblock \emph{Transactions of the Association for Computational Linguistics}, 12:\penalty0 1417--1440, 2024.

\bibitem[Kirk et~al.(2024)Kirk, Mediratta, Nalmpantis, Luketina, Hambro, Grefenstette, and Raileanu]{kirk2024understandingeffectsrlhfllm}
R.~Kirk, I.~Mediratta, C.~Nalmpantis, J.~Luketina, E.~Hambro, E.~Grefenstette, and R.~Raileanu.
\newblock Understanding the effects of rlhf on llm generalisation and diversity, 2024.
\newblock URL \url{https://arxiv.org/abs/2310.06452}.

\bibitem[Kuznetsova et~al.(2017)Kuznetsova, Brockhoff, Christensen, et~al.]{kuznetsova2017lmertest}
A.~Kuznetsova, P.~B. Brockhoff, R.~H. Christensen, et~al.
\newblock lmertest package: tests in linear mixed effects models.
\newblock \emph{Journal of statistical software}, 82\penalty0 (13):\penalty0 1--26, 2017.

\bibitem[Ladosz et~al.(2022)Ladosz, Weng, Kim, and Oh]{ladosz2022exploration}
P.~Ladosz, L.~Weng, M.~Kim, and H.~Oh.
\newblock Exploration in deep reinforcement learning: A survey.
\newblock \emph{Information Fusion}, 85:\penalty0 1--22, 2022.

\bibitem[Lee et~al.(2024{\natexlab{a}})Lee, Dai, Ren, Chen, Cer, Cole, Hui, Boratko, Kapadia, Ding, et~al.]{lee2024gecko}
J.~Lee, Z.~Dai, X.~Ren, B.~Chen, D.~Cer, J.~R. Cole, K.~Hui, M.~Boratko, R.~Kapadia, W.~Ding, et~al.
\newblock Gecko: Versatile text embeddings distilled from large language models.
\newblock \emph{arXiv preprint arXiv:2403.20327}, 2024{\natexlab{a}}.

\bibitem[Lee et~al.(2022)Lee, Liang, and Yang]{10.1145/3491102.3502030}
M.~Lee, P.~Liang, and Q.~Yang.
\newblock Coauthor: Designing a human-ai collaborative writing dataset for exploring language model capabilities.
\newblock In \emph{Proceedings of the 2022 CHI Conference on Human Factors in Computing Systems}, CHI '22, New York, NY, USA, 2022. Association for Computing Machinery.
\newblock ISBN 9781450391573.
\newblock \doi{10.1145/3491102.3502030}.
\newblock URL \url{https://doi.org/10.1145/3491102.3502030}.

\bibitem[Lee et~al.(2024{\natexlab{b}})Lee, Gero, Chung, Shum, Raheja, Shen, Venugopalan, Wambsganss, Zhou, Alghamdi, August, Bhat, Choksi, Dutta, Guo, Hoque, Kim, Knight, Neshaei, Shibani, Shrivastava, Shroff, Sergeyuk, Stark, Sterman, Wang, Bosselut, Buschek, Chang, Chen, Kreminski, Park, Pea, Rho, Shen, and Siangliulue]{10.1145/3613904.3642697}
M.~Lee, K.~I. Gero, J.~J.~Y. Chung, S.~B. Shum, V.~Raheja, H.~Shen, S.~Venugopalan, T.~Wambsganss, D.~Zhou, E.~A. Alghamdi, T.~August, A.~Bhat, M.~Z. Choksi, S.~Dutta, J.~L. Guo, M.~N. Hoque, Y.~Kim, S.~Knight, S.~P. Neshaei, A.~Shibani, D.~Shrivastava, L.~Shroff, A.~Sergeyuk, J.~Stark, S.~Sterman, S.~Wang, A.~Bosselut, D.~Buschek, J.~C. Chang, S.~Chen, M.~Kreminski, J.~Park, R.~Pea, E.~H.~R. Rho, Z.~Shen, and P.~Siangliulue.
\newblock A design space for intelligent and interactive writing assistants.
\newblock In \emph{Proceedings of the 2024 CHI Conference on Human Factors in Computing Systems}, CHI '24, New York, NY, USA, 2024{\natexlab{b}}. Association for Computing Machinery.
\newblock ISBN 9798400703300.
\newblock \doi{10.1145/3613904.3642697}.
\newblock URL \url{https://doi.org/10.1145/3613904.3642697}.

\bibitem[Mirowski et~al.(2023)Mirowski, Mathewson, Pittman, and Evans]{mirowski2023co}
P.~Mirowski, K.~W. Mathewson, J.~Pittman, and R.~Evans.
\newblock Co-writing screenplays and theatre scripts with language models: Evaluation by industry professionals.
\newblock In \emph{Proceedings of the 2023 CHI conference on human factors in computing systems}, pages 1--34, 2023.

\bibitem[Padmakumar and He(2024)]{padmakumar2024doeswritinglanguagemodels}
V.~Padmakumar and H.~He.
\newblock Does writing with language models reduce content diversity?, 2024.
\newblock URL \url{https://arxiv.org/abs/2309.05196}.

\bibitem[Panickssery et~al.(2025)Panickssery, Bowman, and Feng]{panickssery2025llm}
A.~Panickssery, S.~Bowman, and S.~Feng.
\newblock Llm evaluators recognize and favor their own generations.
\newblock \emph{Advances in Neural Information Processing Systems}, 37:\penalty0 68772--68802, 2025.

\bibitem[Park and Lee(2023)]{10.1145/3544548.3581345}
S.~Y. Park and S.~W. Lee.
\newblock Why “why”? the importance of communicating rationales for edits in collaborative writing.
\newblock In \emph{Proceedings of the 2023 CHI Conference on Human Factors in Computing Systems}, CHI '23, New York, NY, USA, 2023. Association for Computing Machinery.
\newblock ISBN 9781450394215.
\newblock \doi{10.1145/3544548.3581345}.
\newblock URL \url{https://doi.org/10.1145/3544548.3581345}.

\bibitem[Raheja et~al.(2023)Raheja, Kumar, Koo, and Kang]{raheja2023coedittexteditingtaskspecific}
V.~Raheja, D.~Kumar, R.~Koo, and D.~Kang.
\newblock Coedit: Text editing by task-specific instruction tuning, 2023.
\newblock URL \url{https://arxiv.org/abs/2305.09857}.

\bibitem[Schick et~al.(2022)Schick, Dwivedi-Yu, Jiang, Petroni, Lewis, Izacard, You, Nalmpantis, Grave, and Riedel]{schick2022peer}
T.~Schick, J.~Dwivedi-Yu, Z.~Jiang, F.~Petroni, P.~Lewis, G.~Izacard, Q.~You, C.~Nalmpantis, E.~Grave, and S.~Riedel.
\newblock Peer: A collaborative language model.
\newblock \emph{arXiv preprint arXiv:2208.11663}, 2022.

\bibitem[Sharma et~al.(2023)Sharma, Tong, Korbak, Duvenaud, Askell, Bowman, Cheng, Durmus, Hatfield-Dodds, Johnston, et~al.]{sharma2023towards}
M.~Sharma, M.~Tong, T.~Korbak, D.~Duvenaud, A.~Askell, S.~R. Bowman, N.~Cheng, E.~Durmus, Z.~Hatfield-Dodds, S.~R. Johnston, et~al.
\newblock Towards understanding sycophancy in language models.
\newblock \emph{arXiv preprint arXiv:2310.13548}, 2023.

\bibitem[Si et~al.(2024{\natexlab{a}})Si, Yang, and Hashimoto]{si2024can}
C.~Si, D.~Yang, and T.~Hashimoto.
\newblock Can llms generate novel research ideas? a large-scale human study with 100+ nlp researchers.
\newblock \emph{arXiv preprint arXiv:2409.04109}, 2024{\natexlab{a}}.

\bibitem[Si et~al.(2024{\natexlab{b}})Si, Yang, and Hashimoto]{si2024llmsgeneratenovelresearch}
C.~Si, D.~Yang, and T.~Hashimoto.
\newblock Can llms generate novel research ideas? a large-scale human study with 100+ nlp researchers, 2024{\natexlab{b}}.
\newblock URL \url{https://arxiv.org/abs/2409.04109}.

\bibitem[Tao et~al.(2024)Tao, Lin, Chen, Li, Wu, Li, Jin, Huang, Tao, and Zhou]{tao2024survey}
Z.~Tao, T.-E. Lin, X.~Chen, H.~Li, Y.~Wu, Y.~Li, Z.~Jin, F.~Huang, D.~Tao, and J.~Zhou.
\newblock A survey on self-evolution of large language models.
\newblock \emph{arXiv preprint arXiv:2404.14387}, 2024.

\bibitem[Thrun(1992)]{thrun1992efficient}
S.~B. Thrun.
\newblock \emph{Efficient exploration in reinforcement learning}.
\newblock Carnegie Mellon University, 1992.

\bibitem[Wang et~al.(2024)Wang, Ma, Feng, Zhang, Yang, Zhang, Chen, Tang, Chen, Lin, Zhao, Wei, and Wen]{Wang_2024}
L.~Wang, C.~Ma, X.~Feng, Z.~Zhang, H.~Yang, J.~Zhang, Z.~Chen, J.~Tang, X.~Chen, Y.~Lin, W.~X. Zhao, Z.~Wei, and J.~Wen.
\newblock A survey on large language model based autonomous agents.
\newblock \emph{Frontiers of Computer Science}, 18\penalty0 (6), Mar. 2024.
\newblock ISSN 2095-2236.
\newblock \doi{10.1007/s11704-024-40231-1}.
\newblock URL \url{http://dx.doi.org/10.1007/s11704-024-40231-1}.

\end{thebibliography}

\end{document}